\documentclass[conference]{IEEEtran}
\IEEEoverridecommandlockouts
\usepackage{cite}
\usepackage{amsmath,amssymb,amsfonts}
\usepackage{algorithm}
\usepackage{algorithmicx}
\usepackage[noend]{algpseudocode}
\usepackage{amsmath}
\usepackage{graphicx}
\usepackage{textcomp}
\usepackage{xcolor}
\usepackage{todonotes}

\def\BibTeX{{\rm B\kern-.05em{\sc i\kern-.025em b}\kern-.08em
    T\kern-.1667em\lower.7ex\hbox{E}\kern-.125emX}}
\begin{document}

\title{Tackling Morpion Solitaire  with AlphaZero-like\\ Ranked Reward Reinforcement Learning\thanks{Hui Wang acknowledges financial support from the China Scholarship Council (CSC), CSC No.201706990015.}
}

\author{\IEEEauthorblockN{Hui Wang, Mike Preuss, Michael Emmerich and Aske Plaat}
\IEEEauthorblockA{Leiden Institute of Advanced Computer Science\\
Leiden University\\
The Netherlands \\
email: h.wang.13@liacs.leidenuniv.nl}}

\maketitle

\begin{abstract}
Morpion Solitaire is a popular single player game, 
performed with paper and pencil. Due to its large state space (on the order of the game of Go) traditional search algorithms, such as MCTS, have not been able to find good solutions. A later algorithm, Nested Rollout Policy Adaptation, was able to find a new record of 82 steps, albeit with large computational resources. After achieving this record, to the best of our knowledge, there has been no further progress reported, for about a decade. 

In this paper we take the recent impressive performance of deep self-learning reinforcement learning approaches from AlphaGo/AlphaZero as inspiration to design a searcher for Morpion Solitaire.  
A challenge of Morpion Solitaire is that the state space is sparse, there are few win/loss signals. Instead, we use an approach known as ranked reward to create a reinforcement learning self-play framework for Morpion Solitaire. This enables us to find medium-quality solutions with reasonable computational effort. Our record is a 67 steps solution, which is very close to the human best (68) without any other adaptation to the problem than using ranked reward. We list many further avenues for potential improvement.
\end{abstract}


\section{Introduction}\label{sec:introduction}
In recent years, the interest in combinatorial games as a challenge in AI has increased after the first AlphaGo program~\cite{silver2016mastering}  defeated the human world champion of Go~\cite{plaat2020learning}. The great success of the AlphaGo and AlphaZero programs~\cite{silver2016mastering,silver2017mastering,silver2018general} in two-player games, has inspired attempts in other domains~\cite{segler2018planning,wang2020warm}. So far, one of the most challenging single player games, Morpion Solitaire~\cite{cboyer2020} has not yet been studied with this promising deep reinforcement learning approach. 

Morpion Solitaire is a popular single player game since 1960s~\cite{cboyer2020,demaine2006morpion}, because of its simple rules and simple equipment, requiring only  paper and  pencil. Due to its large state space it is also an interesting AI challenge in single player games, just like the game of Go challenge in two-player turn-based games. Could the AlphaZero self-play approach, so successful in Go, also work in Morpion Solitaire? For ten years little progress has been made in Morpion Solitaire. It is time to take up the challenge and to see if a self-play deep reinforcement learning approach will work in this challenging game. 

AlphaGo and AlphaZero combine deep neural networks~\cite{schmidhuber2015deep} and Monte Carlo Tree Search~(MCTS)~\cite{browne2012survey} in a self-play framework that learns by curriculum learning~\cite{bengio2009curriculum}. Unfortunately, these approaches can not be directly used to play single agent combinatorial games, such as travelling salesman problems~(TSP)~\cite{rego2011traveling} and bin package problems~(BPP)~\cite{hu2018multi}, where cost minimization is the goal of the game. To apply self-play for single player games, Laterre et al. proposed a Ranked Reward~(R2) algorithm. R2 creates a relative performance metric by means of ranking the rewards obtained by a single agent over multiple games. In two-dimensional and three-dimensional bin packing R2 is reported to out-perform MCTS~\cite{laterre2018ranked}. In this paper we use this idea for Morpion Solitaire. Our contributions can be summarized as follows:
\begin{enumerate}
    \item We present the first implementation\footnote{Source code: https://github.com/wh1992v/R2RRMopionSolitaire} of Ranked Reward AlphaZero-style self-play for Morpion Solitaire.
    \item On this implementation, we report our current best solution, of 67 steps~(see Fig~\ref{fig:67stepsgrid}). 
\end{enumerate} 
This result is very close to the human record, and shows the potential of the self-play reinforcement learning approach in Morpion Solitaire, and other hard single player combinatorial problems.

This paper is structured as follows. 
After giving an overview of related work in Sect.\,\ref{sec:relatedwork}, we introduce the 
Morpion Solitaire challenge in Sect.\,\ref{sec:morpion}. Then we present how to integrate the idea of R2 into  AlphaZero self-play in Sect.\,\ref{sec:R2}. Thereafter, we set up the experiment in Sect.\,\ref{sec:setup}, and show the result and analysis in Sect.\,\ref{sec:result}. Finally, we conclude our paper and discuss future work.

\section{Related Work}\label{sec:relatedwork}
Deep reinforcement learning~\cite{mnih2015human} approaches, especially the AlphaGo and AlphaZero programs, which combine online tree search and offline neural network training, achieve super human level of playing two player turn-based board games such as Go, Chess and Shogi~\cite{silver2016mastering,silver2017mastering,silver2018general}. These  successes spark the interest of creating new deep reinforcement learning approaches to solve  problems in the field of game AI, especially for other two player games~\cite{tian2019elf,wang2020analysis,wang2018monte,wang2018assessing}.

However, for single player games, self-play deep reinforcement learning approaches are not yet well studied since the approaches used for two-player games can not directly be used in single player games~\cite{laterre2018ranked}, since the goal of the task changes from winning from an opponent, to minimizing the solution cost. Nevertheless, some researchers did initial works on single games with self-play deep reinforcement learning~\cite{moerland2018a0c}. The main difficulty is representing single player games in ways that allow the use of a deep reinforcement learning approach. In order to solve this difficulty, Vinyals et al.~\cite{vinyals2015pointer} proposed a neural architecture~(Pointer Networks) to represent combinatorial optimization problems as sequence-to-sequence
learning problems. Early Pointer Networks achieved decent performance on TSP, but this approach is computationally expensive and  requires handcrafted training examples for supervised learning methods. Replacing supervised learning methods by actor-critic methods removed this requirement~\cite{bello2016neural}. In addition, Laterre et al.\ proposed the R2 algorithm through ranking the rewards obtained by a single agent over multiple games to label win or loss for each search, and this algorithm reportedly outperformed plain MCTS in the bin packing problem (BPP)~\cite{laterre2018ranked}. Feng et al.\ recently used curriculum-driven deep reinforcement learning to solve hard Sokoban instances \cite{feng2020solving}.

In addition to TSP and BPP, Morpion Solitaire has long been a  challenge in NP-hard single player problems. Previous works on Morpion Solitaire mainly employ traditional heuristic search algorithms~\cite{cboyer2020}. Cazenave created Nested Monte-Carlo Search and found an 80 moves record~\cite{cazenave2009nested}. After that, a new Nested Rollout Policy Adaptation algorithm achieved a new 82 steps record~\cite{rosin2011nested}. Thereafter, Cazenave applied Beam Nested Rollout Policy Adaptation~\cite{cazenave2012beam}, which reached the same 82 steps record but did not exceed it, indicating the difficulty of making further progress on Morpion Solitaire using traditional search heuristics.

We believe that it is time for a new approach, applying (self-play) deep reinforcement learning  to train a Morpion Solitaire player. The combination of the R2 algorithm with the AlphaZero self-play framework could be a first alternative for above mentioned approaches.

\section{Morpion Solitaire}\label{sec:morpion}
\begin{figure}[htbp]
\centerline{\includegraphics[width=0.9\columnwidth]{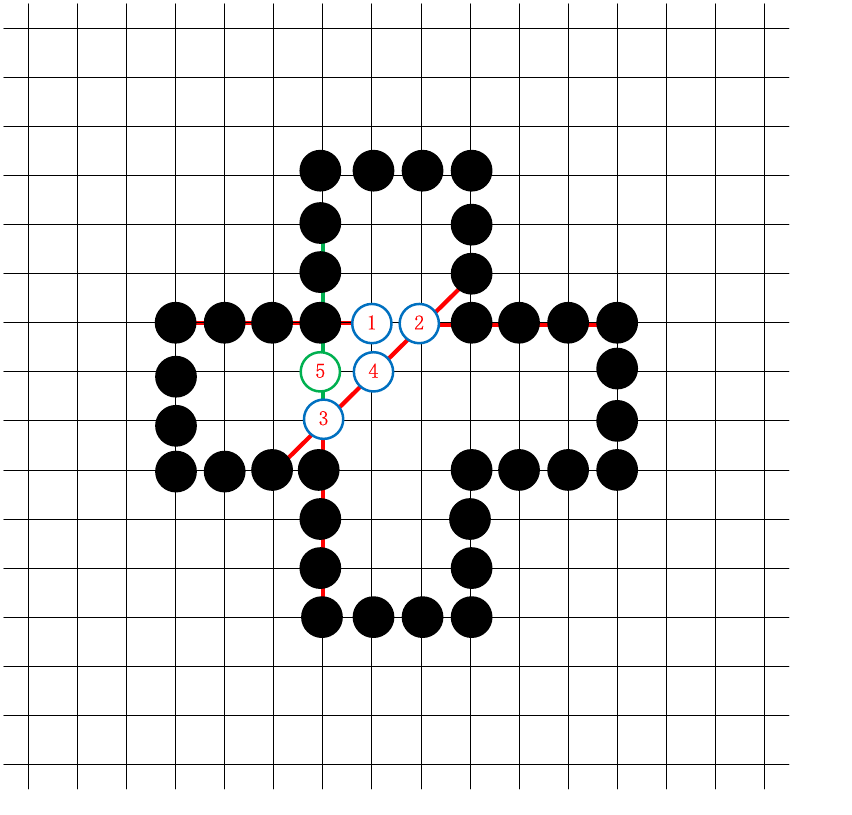}}
\caption{Moves Example: Moves 1, 2, 3, 4 are legal moves, move 5 is illegal for the 5D version, but legal for the 5T version.}
\label{fig:ruleexample}
\end{figure}

Morpion Solitaire is a single player game played on an unlimited grid. It is a well know NP-hard challenge~\cite{cboyer2020}. The rules of the game are simple. There are 36 black circles as the initial state~(see Fig~\ref{fig:ruleexample}). A move for Morpion Solitaire consists of two parts: a) placing a new circle on the paper so that this new circle can be connected with four other existing circles horizontally, vertically or diagonally, and then b) drawing a line to connect these five circles~(see action 1, 2, 3 in the figure). A line is allowed to cross over each other~(action 4), but not 
allowed to overlap. There are two versions: the Touching~(5T) version and the Disjoint~(5D) version. For the 5T version, it is allowed to touch~(action 5, green circle and green line), but for the 5D version, touching is illegal~(any circle can not belong to two lines that have the same direction). After  a legal action the circle and the line are added to the grid. In this paper we are interested in the 5D version.

The best human
score for the 5D version is 68 moves \cite{demaine2006morpion}. A score of 80 moves was found by means of Nested Monte-Carlo search~\cite{cazenave2009nested}. In addition,~\cite{rosin2011nested} found a new record with 82 steps, and \cite{cazenave2012beam} also found a 82 steps solution. It has been proven mathematically  that the 5D version has an upper bound of 121~\cite{kawamura2013morpion}. 

\section{Ranked Reward Reinforcement Learning}\label{sec:R2}
AlphaZero self-play achieved milestone successes in  two-player games, but  can not be directly used for single player cost minimization games. Therefore, the R2 algorithm has been  created to use self-play
for generic single player MDPs. R2 reshapes the rewards according to player's relative performance over recent games~\cite{laterre2018ranked}. The pseudo code of  R2  is given in Algorithm~\ref{alg:r2a0g}.

\begin{algorithm*}[bth!]
\caption{Ranked Reward Reinforcement Learning within AlphaZero-like Self-play Framework}
\label{alg:r2a0g}
\begin{algorithmic}[1]
\footnotesize
\Function{RankedRewardReinforcementLearning}{}
\State Initialize $f_\theta$ with random weights; Initialize retrain buffer $D$ and  reward list $B$ 
\For{iteration=1, $\dots$ $\dots$, $I$}
\Comment self-play curriculum of $I$ tournaments
\For{episode=1,$\dots$, $E$}\Comment{stage 1, 
self-play tournament of $E$ games}
\For{t=1, $\dots$, $T^\prime$, $\dots$, $T$} 
\Comment play game of $T$ moves

\State$\pi_{t} \leftarrow$ perform MCTS based on  $f_\theta$ or directly get policy from  $f_\theta$
\Comment with or without tree search enhancement \label{mctsenhancement}
\State $a_t=$randomly select on $\pi_t$ before $T^\prime$ or $\arg\max_a(\pi_t)$ after $T^\prime$ step
\State executeAction($s_t$, $a_t$)
\EndFor

\State Calculate game reward $r_{T}$ and store it in $B$\label{line:calreward}
\State Calculate threshold $r_\alpha$ based on the recent games rewards in $B$

\State Reshape the ranked reward $z$ following Equation~\ref{equ:reshape}\label{line:reshape}\Comment{Ranked Reward}

\State Store every $(s_t,\pi_t,z_t)$ with ranked rewards $z_t$ ~($t\in[1,T]$) in $D$
\EndFor
\State Randomly sample  minibatch of examples~($s_j$, $\pi_j$, $z_j$) from $D$ \Comment{stage 2}
\State Train $f_{\theta^\prime}\leftarrow f_\theta$ 
\State Play once with MCTS enhancement on $f_{\theta^\prime}$\label{mctsplay} \Comment{stage 3}
\State Replace $f_{\theta}\leftarrow f_\theta^\prime$ 

\EndFor
\State \Return $f_\theta$;
\EndFunction
\end{algorithmic}
\end{algorithm*}

Following  AlphaZero-like self-play~\cite{wang2019alternative}, we demonstrate the typical three stages as shown in the pseudo code. For self-play in Morpion Solitaire MCTS  is too time consuming due to the large state space. Thus, we rely on the policy directly from $f_\theta$ without tree search~(line~\ref{mctsenhancement}). For stage 3, we directly replace the previous neural network model with the newly trained model. and we let the newly trained model play a single time with MCTS enhancement~(line~\ref{mctsplay}).  The R2 idea is integrated~(see line~\ref{line:calreward} to line~\ref{line:reshape}). The reward list $B$ stores the recent game rewards. According to a ratio $\alpha$, the threshold of $r_\alpha$ is calculated. We then compare  $r_\alpha$ to the game reward $r_T$ to reshape the ranked reward $z$ according to Equation~\ref{equ:reshape}.

\begin{equation}\label{equ:reshape}
    z=
\begin{cases}
1& r_T > r_\alpha\\
-1& r_T < r_\alpha\\
random(1, -1) & r_T = r_\alpha
\end{cases}
\end{equation}
where $r_\alpha$ is the stored reward value in $B$ indexed by $L\times \alpha$, $L$ is the length of $B$, $\alpha$ is a ratio parameter. 

\section{Experiment Setup}\label{sec:setup}
We perform our experiments on a GPU server with 128G RAM, 3TB local storage, 20 Intel Xeon E5-2650v3 cores~(2.30GHz, 40 threads), 2 NVIDIA Titanium GPUs~(each with 12GB memory) and 6 NVIDIA GTX 980 Ti GPUs~(each with 6GB memory). 

The hyper-parameters of our current R2 implementation are as much as possible equal to previous work. In this work, all neural network models share the same structure as in~\cite{wang2019alternative}. The hyper-parameter values for Algorithm~\ref{alg:r2a0g} used in our experiments are given in Table~\ref{defaulttab}. Partly, these  values are set based on the work reported in~\cite{wang2019hyper} and the R2 approach for BPP~\cite{laterre2018ranked}. $T'$ is set to half of the current best record. $m$ is set to 100 if using MCTS in self-play, but 20000 for MCTS in stage 3. Furthermore, as there is an upper bound of the best score~(121), we did experiments on 16$\times$16, 20$\times$20 and 22$\times$22 boards respectively. Training time for every algorithm is about a week.

\begin{table}[!bth]
\small
\centering
\caption{Default Parameter Settings}\smallskip
\resizebox{1\columnwidth}{!}{
\smallskip\begin{tabular}{|c|c|c|}
\hline
Parameter	& Brief Description& Default Value\\
\hline
\emph{I}	&number of iterations 	&100	\\
\hline
\emph{E}    &number of episodes   &50\\
\hline
\emph{T'}	&step threshold	&41\\
\hline
\emph{m}	&MCTS simulation times &20000\\
\hline
\emph{c}    &weight in UCT  &1.0\\
\hline
\emph{rs}	& number of retrain iterations &10\\
\hline
\emph{ep}	& number of epochs	&5	\\
\hline
\emph{bs}	& batch size	&64\\
\hline
\emph{lr}	& learning rate	&0.005\\
\hline
\emph{d}    & dropout probability  &0.3\\
\hline
\emph{L}    & length of $B$ &200\\
\hline
\emph{$\alpha$}	& ratio to compute $r_\alpha$	&0.75 \\
\hline
\end{tabular}
}
\label{defaulttab}
\end{table}

\section{Result and analysis}\label{sec:result}

\begin{figure}[htbp]
\centerline{\includegraphics[width=0.9\columnwidth]{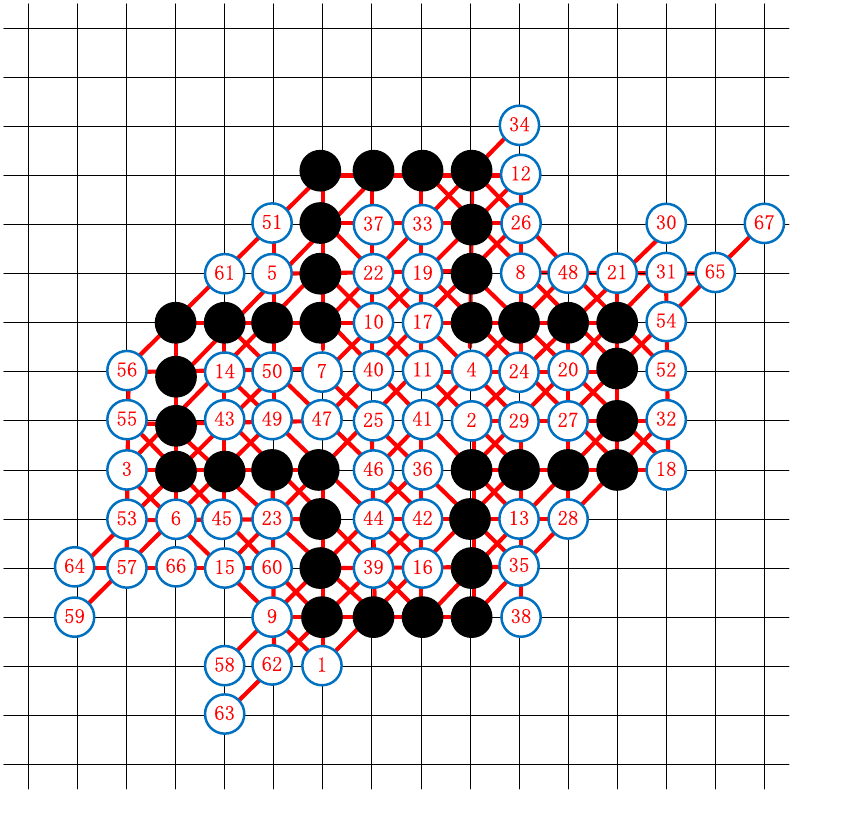}}
\caption{Detailed Steps of Our Best Solution}
\label{fig:67stepsgrid}
\end{figure}

As we mentioned above, the best score for Morpion Solitaire of 82 steps has been achieved by Nested Rollout Policy Adaptation in 2010. The best score achieved by human is 68. Our first attempt with limited computation resources on a large size board~(22$\times$22) achieved a score of 67, very close to the best human score. The resulting solution is  shown in Fig~\ref{fig:67stepsgrid}.

Based on these promising results with  Ranked Reward Reinforcement Learning we identify  areas for further improvement. First,  parameter values for the Morpion Solitaire game can be fine-tuned using results of small board games. Especially the parameter  $m=100$ seems not sufficient for large boards. Second, the neural network could be changed to Pointer Networks and the size of neural network should be deeper.  

Note that the tuning of parameters is critical; if the reward list $B$ is too small,  the reward list can be easily filled up by scores close to 67. The training will then be stuck in a locally optimal solution. As good solutions are expected to be sparsely distributed over the  search space, this increases the difficulty to get rid of a locally optimal solution once the algorithm has focused on it.

\section{Conclusion and Outlook}\label{sec:conclusion}
In this work, we apply a Ranked Reward Reinforcement Learning AlphaZero-like approach to play Morpion Solitaire,  an important NP-hard  single player game challenge. We train the player on 16$\times$16, 20$\times$20 and 22$\times$22 boards, and find a near best human performance solution with 67 steps.  As a first attempt of utilizing self-play deep reinforcement learning approach to tackle Morpion Solitaire, achieving near-human performance is a promising result.  

Our first results give us reason to believe that there remain ample possibilities to improve the approach by investigating the following aspects:
\begin{enumerate}
    \item Parameter tuning: such as the Monte Carlo simulation times. Since good solutions are  sparse in this game, maybe more exploration is beneficial?
    \item Neural Network Design: It is reported that Pointer Networks perform better on combinatorial problems. A next step could be to also make the neural network structure deeper.
    \item Local Optima: Through monitoring the reward list $B$, we can adjust in time by enlarging more exploration once it gets stuck in a locally optimal solution. 
    \item Computation resources and parallelization: enhanced parallelization may improve the results.
\end{enumerate}

To summarize, although the problem is difficult due to its large state space and sparsity of good solutions, applying a Ranked Reward self-play Reinforcement Learning approach to tackle Morpion Solitaire is a promising and learns from \textit{tabula rasa}.  We present our promising near-human result to stimulate future work on Morpion Solitaire and other single agent games with self-play reinforcement learning. 

\section*{Acknowledgment}

Hui Wang acknowledges financial support from the China Scholarship Council (CSC), CSC No.201706990015.

\bibliographystyle{./IEEEtran}
\bibliography{.IEEEabrv,./IEEEexample}

\begin{thebibliography}{10}
\providecommand{\url}[1]{#1}
\csname url@samestyle\endcsname
\providecommand{\newblock}{\relax}
\providecommand{\bibinfo}[2]{#2}
\providecommand{\BIBentrySTDinterwordspacing}{\spaceskip=0pt\relax}
\providecommand{\BIBentryALTinterwordstretchfactor}{4}
\providecommand{\BIBentryALTinterwordspacing}{\spaceskip=\fontdimen2\font plus
\BIBentryALTinterwordstretchfactor\fontdimen3\font minus
  \fontdimen4\font\relax}
\providecommand{\BIBforeignlanguage}[2]{{%
\expandafter\ifx\csname l@#1\endcsname\relax
\typeout{** WARNING: IEEEtran.bst: No hyphenation pattern has been}%
\typeout{** loaded for the language `#1'. Using the pattern for}%
\typeout{** the default language instead.}%
\else
\language=\csname l@#1\endcsname
\fi
#2}}
\providecommand{\BIBdecl}{\relax}
\BIBdecl

\bibitem{silver2016mastering}
D.~Silver, A.~Huang, C.~J. Maddison, A.~Guez, L.~Sifre, G.~Van Den~Driessche,
  J.~Schrittwieser, I.~Antonoglou, V.~Panneershelvam, M.~Lanctot \emph{et~al.},
  ``Mastering the game of go with deep neural networks and tree search,''
  \emph{nature}, vol. 529, no. 7587, p. 484, 2016.

\bibitem{plaat2020learning}
A.~Plaat, \emph{Learning to Play: Reinforcement Learning and Games}.\hskip 1em
  plus 0.5em minus 0.4em\relax Springer Verlag, Heidelberg, New York, 2020.

\bibitem{silver2017mastering}
D.~Silver, J.~Schrittwieser, K.~Simonyan, I.~Antonoglou, A.~Huang, A.~Guez,
  T.~Hubert, L.~Baker, M.~Lai, A.~Bolton \emph{et~al.}, ``Mastering the game of
  go without human knowledge,'' \emph{Nature}, vol. 550, no. 7676, p. 354,
  2017.

\bibitem{silver2018general}
D.~Silver, T.~Hubert, J.~Schrittwieser, I.~Antonoglou, M.~Lai, A.~Guez,
  M.~Lanctot, L.~Sifre, D.~Kumaran, T.~Graepel \emph{et~al.}, ``A general
  reinforcement learning algorithm that masters chess, shogi, and go through
  self-play,'' \emph{Science}, vol. 362, no. 6419, pp. 1140--1144, 2018.

\bibitem{segler2018planning}
M.~H. Segler, M.~Preuss, and M.~P. Waller, ``Planning chemical syntheses with
  deep neural networks and symbolic ai,'' \emph{Nature}, vol. 555, no. 7698,
  pp. 604--610, 2018.

\bibitem{wang2020warm}
H.~Wang, M.~Preuss, and A.~Plaat, ``Warm-start alphazero self-play search
  enhancements,'' \emph{arXiv preprint arXiv:2004.12357}, 2020.

\bibitem{cboyer2020}
C.~Boyer, ``Morpion solitaire,'' \url{http://www.morpionsolitaire.com/}, 2020,
  accessed May, 2020.

\bibitem{demaine2006morpion}
E.~D. Demaine, M.~L. Demaine, A.~Langerman, and S.~Langerman, ``Morpion
  solitaire,'' \emph{Theory of Computing Systems}, vol.~39, no.~3, pp.
  439--453, 2006.

\bibitem{schmidhuber2015deep}
J.~Schmidhuber, ``Deep learning in neural networks: An overview,'' \emph{Neural
  networks}, vol.~61, pp. 85--117, 2015.

\bibitem{browne2012survey}
C.~B. Browne, E.~Powley, D.~Whitehouse, S.~M. Lucas, P.~I. Cowling,
  P.~Rohlfshagen, S.~Tavener, D.~Perez, S.~Samothrakis, and S.~Colton, ``A
  survey of monte carlo tree search methods,'' \emph{IEEE Transactions on
  Computational Intelligence and AI in games}, vol.~4, no.~1, pp. 1--43, 2012.

\bibitem{bengio2009curriculum}
Y.~Bengio, J.~Louradour, R.~Collobert, and J.~Weston, ``Curriculum learning,''
  in \emph{Proceedings of the 26th annual international conference on machine
  learning}.\hskip 1em plus 0.5em minus 0.4em\relax ACM, 2009, pp. 41--48.

\bibitem{rego2011traveling}
C.~Rego, D.~Gamboa, F.~Glover, and C.~Osterman, ``Traveling salesman problem
  heuristics: Leading methods, implementations and latest advances,''
  \emph{European Journal of Operational Research}, vol. 211, no.~3, pp.
  427--441, 2011.

\bibitem{hu2018multi}
H.~Hu, L.~Duan, X.~Zhang, Y.~Xu, and J.~Wei, ``A multi-task selected learning
  approach for solving new type 3d bin packing problem,'' \emph{arXiv preprint
  arXiv:1804.06896}, 2018.

\bibitem{laterre2018ranked}
A.~Laterre, Y.~Fu, M.~K. Jabri, A.-S. Cohen, D.~Kas, K.~Hajjar, T.~S. Dahl,
  A.~Kerkeni, and K.~Beguir, ``Ranked reward: Enabling self-play reinforcement
  learning for combinatorial optimization,'' \emph{arXiv preprint
  arXiv:1807.01672}, 2018.

\bibitem{mnih2015human}
V.~Mnih, K.~Kavukcuoglu, D.~Silver, A.~A. Rusu, J.~Veness, M.~G. Bellemare,
  A.~Graves, M.~Riedmiller, A.~K. Fidjeland, G.~Ostrovski \emph{et~al.},
  ``Human-level control through deep reinforcement learning,'' \emph{Nature},
  vol. 518, no. 7540, pp. 529--533, 2015.

\bibitem{tian2019elf}
Y.~Tian, J.~Ma, Q.~Gong, S.~Sengupta, Z.~Chen, J.~Pinkerton, and C.~L. Zitnick,
  ``Elf opengo: An analysis and open reimplementation of alphazero,''
  \emph{arXiv preprint arXiv:1902.04522}, 2019.

\bibitem{wang2020analysis}
H.~Wang, M.~Emmerich, M.~Preuss, and A.~Plaat, ``Analysis of hyper-parameters
  for small games: Iterations or epochs in self-play?'' \emph{arXiv preprint
  arXiv:2003.05988}, 2020.

\bibitem{wang2018monte}
H.~Wang, M.~Emmerich, and A.~Plaat, ``Monte carlo q-learning for general game
  playing,'' \emph{arXiv preprint arXiv:1802.05944}, 2018.

\bibitem{wang2018assessing}
------, ``Assessing the potential of classical q-learning in general game
  playing,'' in \emph{Benelux Conference on Artificial Intelligence}.\hskip 1em
  plus 0.5em minus 0.4em\relax Springer, 2018, pp. 138--150.

\bibitem{moerland2018a0c}
T.~M. Moerland, J.~Broekens, A.~Plaat, and C.~M. Jonker, ``A0c: Alpha zero in
  continuous action space,'' \emph{arXiv preprint arXiv:1805.09613}, 2018.

\bibitem{vinyals2015pointer}
O.~Vinyals, M.~Fortunato, and N.~Jaitly, ``Pointer networks,'' in
  \emph{Advances in neural information processing systems}, 2015, pp.
  2692--2700.

\bibitem{bello2016neural}
I.~Bello, H.~Pham, Q.~V. Le, M.~Norouzi, and S.~Bengio, ``Neural combinatorial
  optimization with reinforcement learning,'' \emph{arXiv preprint
  arXiv:1611.09940}, 2016.

\bibitem{feng2020solving}
D.~Feng, C.~P. Gomes, and B.~Selman, ``Solving hard ai planning instances using
  curriculum-driven deep reinforcement learning,'' \emph{arXiv preprint
  arXiv:2006.02689}, 2020.

\bibitem{cazenave2009nested}
T.~Cazenave, ``Nested monte-carlo search,'' in \emph{Twenty-First International
  Joint Conference on Artificial Intelligence}, 2009.

\bibitem{rosin2011nested}
C.~D. Rosin, ``Nested rollout policy adaptation for monte carlo tree search,''
  in \emph{Twenty-Second International Joint Conference on Artificial
  Intelligence}, 2011.

\bibitem{cazenave2012beam}
T.~Cazenave and F.~Teytaud, ``Beam nested rollout policy adaptation,'' 2012.

\bibitem{kawamura2013morpion}
A.~Kawamura, T.~Okamoto, Y.~Tatsu, Y.~Uno, and M.~Yamato, ``Morpion solitaire
  5d: a new upper bound of 121 on the maximum score,'' \emph{arXiv preprint
  arXiv:1307.8192}, 2013.

\bibitem{wang2019alternative}
H.~Wang, M.~Emmerich, M.~Preuss, and A.~Plaat, ``Alternative loss functions in
  alphazero-like self-play,'' in \emph{2019 IEEE Symposium Series on
  Computational Intelligence (SSCI)}.\hskip 1em plus 0.5em minus 0.4em\relax
  IEEE, 2019, pp. 155--162.

\bibitem{wang2019hyper}
------, ``Hyper-parameter sweep on alphazero general,'' \emph{arXiv preprint
  arXiv:1903.08129}, 2019.

\end{thebibliography}

\end{document}